\documentclass{article}
\usepackage{spconf,amsmath,epsfig}

\let\OLDthebibliography\thebibliography
\renewcommand\thebibliography[1]{
  \OLDthebibliography{#1}
  \setlength{\parskip}{0pt}
  \setlength{\itemsep}{0pt plus 0.3ex}
}

\begin{document}\sloppy

\def\x{{\mathbf x}}
\def\L{{\cal L}}

\title{A Survey on Backbones for Deep Video Action Recognition}
\name{Zixuan Tang\textsuperscript{1}, Youjun Zhao\textsuperscript{1}, Yuhang Wen\textsuperscript{1}, Mengyuan Liu\textsuperscript{2}* \thanks{* Corresponding author. Email: nkliuyifang@gmail.com}}

\address{\textsuperscript{1}School of Intelligent Systems Engineering,\\
Sun Yat-sen University, Shenzhen 518107, China\\
         \textsuperscript{2}National Key Laboratory of General Artificial Intelligence,\\
         Peking University, Shenzhen Graduate School, Shenzhen 518055, China}

\maketitle

\begin{abstract}
Action recognition is a key technology in building interactive metaverses. With the rapid development of deep learning, methods in action recognition have also achieved great advancement. Researchers design and implement the backbones referring to multiple standpoints, which leads to the diversity of methods and encountering new challenges. This paper reviews several action recognition methods based on deep neural networks. We introduce these methods in three parts: 1) Two-Streams networks and their variants, which, specifically in this paper, use RGB video frame and optical flow modality as input; 2) 3D convolutional networks, which make efforts in taking advantage of RGB modality directly while extracting different motion information is no longer necessary; 3) Transformer-based methods, which introduce the model from natural language processing into computer vision and video understanding. We offer objective sights in this review and hopefully provide a reference for future research.
\end{abstract}
\section{Introduction}
Video action recognition is a foundational technology for building the metaverse because it meets the needs for immersive interactive experiences~\cite{mystakidis2022metaverse}. With the development of deep learning and computing power, deep neural network gradually takes a dominant place in computer vision. Convolutional Neural Network (CNN) was primarily designed for image classification. Due to its great success in the image domain, CNN-based methods are extended into video understanding. Besides, the Transformer was introduced into computer vision and achieved great success. This result leads to the research concerning Transformer-based methods in video understanding and action recognition.
CNN has shown remarkable performance in tasks related to still images, such as image classification and semantic segmentation. 
Two-streams Networks method~\cite{simonyan2014two} introduces an additional pathway that takes optical flow as input, which indicates the temporal information in videos. The success of Two-Streams networks inspired many follow-up researches~\cite{Wang_2015_CVPR,wang2016temporal}. Another CNN method fuses the temporal information with spatial features by directly using a 3D convolutional filter~\cite{tran2015learning}. The research in 3D convolutional networks has also achieved great advancement~\cite{carreira2017quo,hara2018can,feichtenhofer2019slowfast,feichtenhofer2020x3d}. 
Transformer is proven to be also effective in computer vision without the convolutional filter. Its attention mechanism allows the model to perform video action recognition more accurately in metaverse scenarios. In recent years there are many Transformer-based methods being proposed~\cite{bertasius2021space,neimark2021video,arnab2021vivit,fan2021multiscale,liu2021video,wu2022memvit}. 
The rapid development of video action recognition comes with various methods and complex challenges, which inspire us to provide this review to offer new sights in this area. In this paper:
\textbf{1)} We overview the development of video action recognition and introduce popular benchmarks in video action recognition.
\textbf{2)} We review several notable deep neural network methods in video action recognition, which include Transformer-based networks. 
\textbf{3)} We make an inventory summary of future challenges and promising directions in the video action recognition field.





\begin{figure}[t]
    \begin{center}
    \includegraphics[width=0.4\textwidth]{ 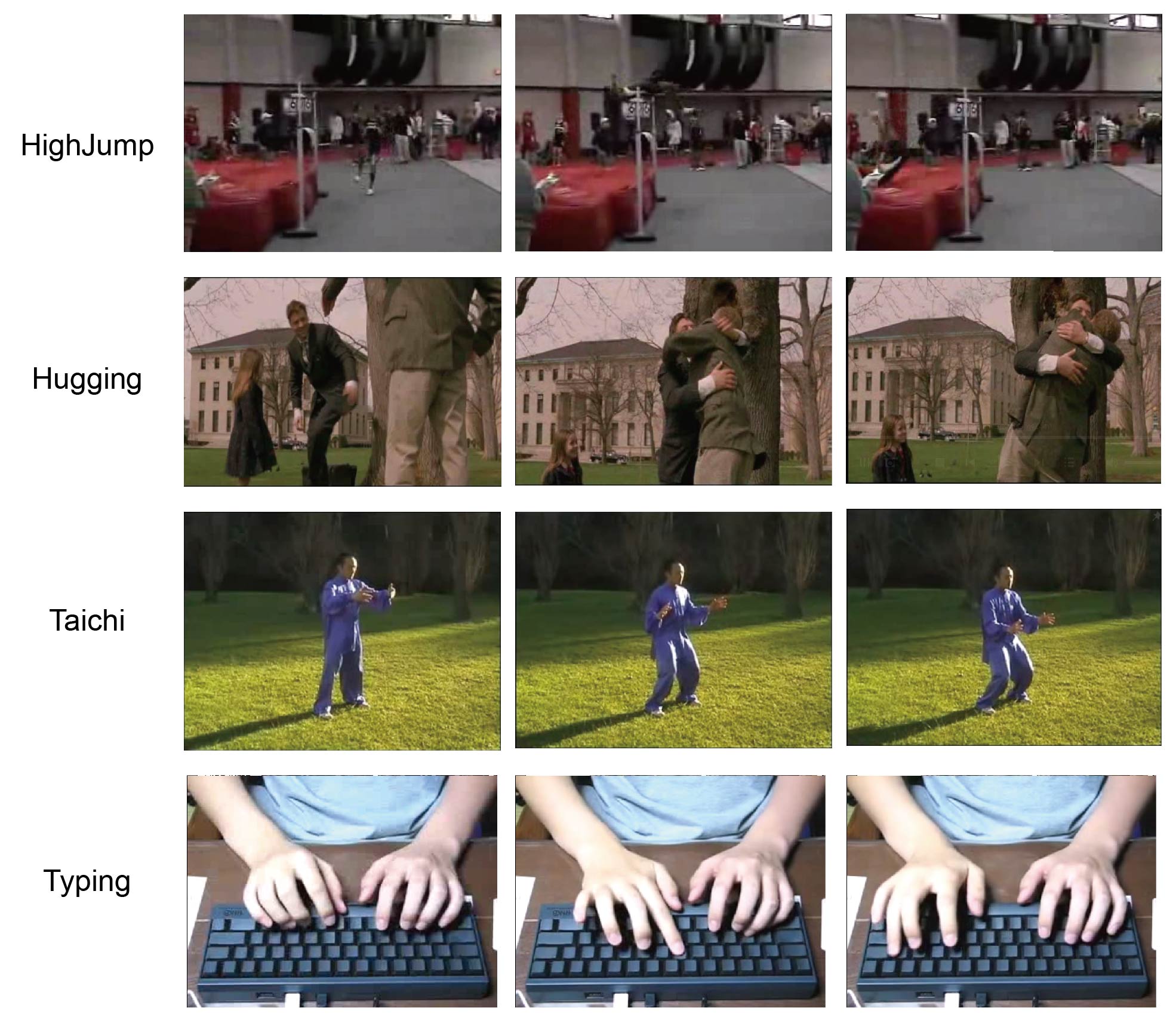}    
    \end{center}\vspace{-1em}
    \caption{Over the last decade, many video action recognition datasets with various labels have been proposed, which contributes to the advancement of action recognition tasks.}
    \label{datasample}
\end{figure}

\section{Backbones of Action Recognition}

\begin{figure*}[t]
    \begin{center}
    \includegraphics[width=0.9\textwidth]{  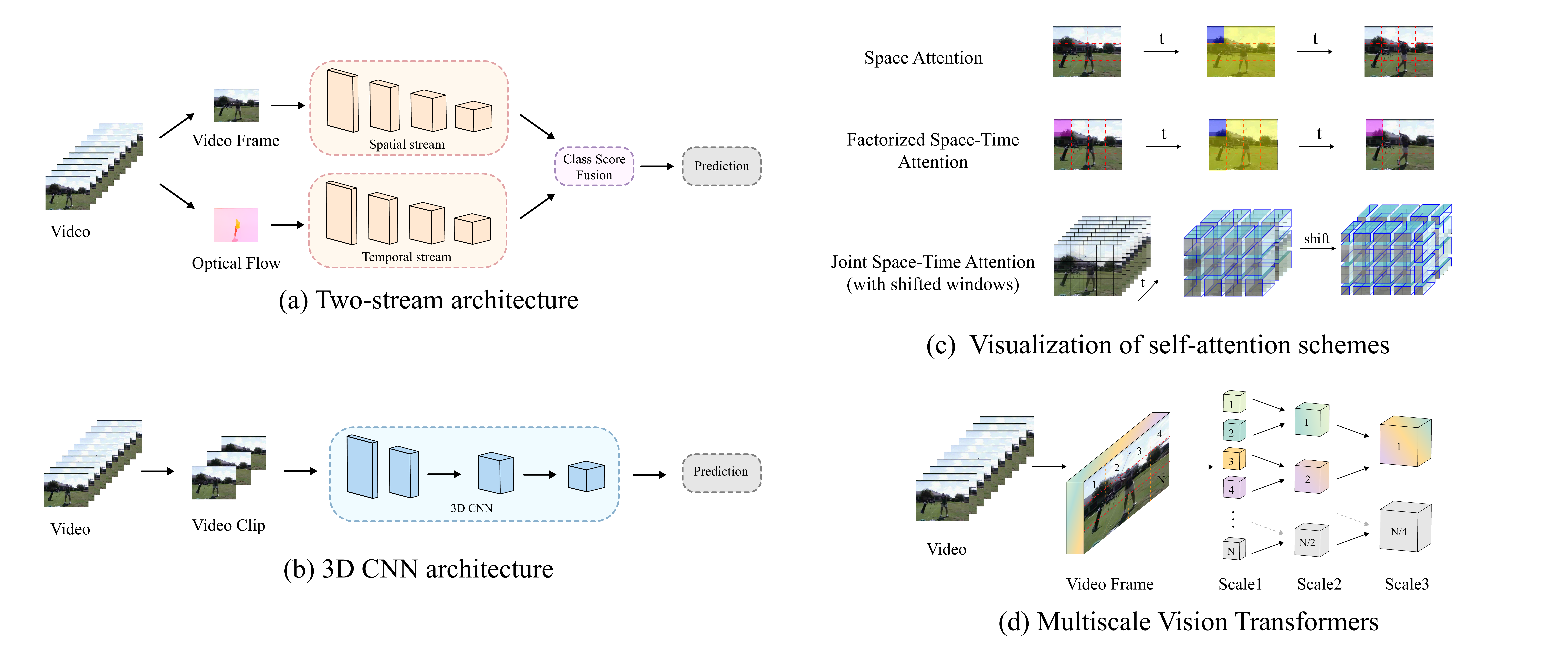}    
    \end{center}\vspace{-1em}
    \caption{We review deep neural network backbones in video action recognition. As shown in Fig. \ref{network}, we demonstrate the general architecture of (a) Two-stream networks and (b) 3D CNN. Moreover, we review two ways of improving Transformer for action recognition: designing different kinds of attention mechanisms (c) or introducing multi-scale/multi-view features(d) into the model.}
    \label{network}
\end{figure*}


\subsection{Two-Streams Networks in Action Recognition}

\subsubsection{Two-Stream Networks}Two-stream network based on a single-stream network is developed by using both spatial stream and temporal stream as input to extract video information. Two-Stream Neural Networks~\cite{simonyan2014two} was proposed in 2014, which expanded CNN in the image domain into the spatial stream and temporal stream. As shown in Fig. \ref{network}(a), the spatial stream is the same as the CNN in the image domain, which performs action recognition in space. The temporal stream takes a stacked optical flow, which represents temporal components from videos. After the parallel process of two streams, the inference results from two CNNs were fused into the class score. This fusion method is referred to as late fusion~\cite{karpathy2014large}, which means no information interaction during the feature extraction. 

An additional temporal stream makes it possible for CNN to achieve performance equal to traditional handcrafted features and inspired many follow-up researchers. TDD~\cite{Wang_2015_CVPR} proposes to improve the performance of CNN with trajectory-constrained pooling, which merits benefit from both deep network and handcrafted features. 
Then TSN ~\cite{wang2016temporal} proposes to model long-range temporal information by sparse sampling.

\subsubsection{Multi-Stream Networks}


Action recognition tasks can perform well with pose estimation information. P-CNN~\cite{cheron2015p} aggregated the appearance and motion information of human pose by tracking the human parts. In particular, it used the position of joints to represent body regions with CNN descriptors to conduct the P-CNN feature. PoTion~\cite{choutas2018potion} utilized the movement of human joints over the video clips as semantic key points to represent human action. ~\cite{liu2019joint} propose dynamic pose image (DPI) as a compact pose feature for human action recognition. Based on joint estimation maps, DPI captures richer information about human body parts, compared with pose-based methods using joint locations. ~\cite{liu2016salient} introduces salient directed graphs with Time Salient and Space Salient Pairwise features for efficient real-time human action recognition.
Object information can also benefit action recognition. 
R*CNN~\cite{gkioxari2015contextual} adapted RCNN to predict action in more than one region. It observed that action videos are always combined with contextual cues like objects and scenes, providing an additional source of information for video understanding. 


\subsection{3D CNNs}

\subsubsection{Inspiration from Image Domain}

The general pipeline for 3D CNN is shown in Fig. \ref{network}(b), which is much like to image domain. One of the most challenging issues of using 3D convolution is that the parameters make the training and convergence of the model harder. In 2017, Carreira \textit{et al.} ~\cite{carreira2017quo} proposed the I3D model, whose architecture is inspired by Inception ~\cite{szegedy2015going} and can initiate by inflating the 2D convolution of the Inception network to 3D convolution. The inflating technique makes it possible for 3D CNN to utilize existing large-scale image datasets like ImageNet. The newly proposed larger Kinetics dataset was also used in the training process of the I3D network, which played an important role in the convergence of the network. The I3D network achieves better accuracy than the 3D Convolution technique proposed before~\cite{ji20123d,tran2015learning} on the HMDB-51 datasets using only RGB as input. 


\subsubsection{Spatiotemporal Semantic Information}

Highlighting the significance of spatio-temporal information, various approaches have been proposed to analyze video features. Qiu \textit{et al.} ~\cite{qiu2017learning} introduced the P3D model, which simplifies optimization by separating spatial and temporal convolution parameters. Similarly, Tran \textit{et al.} ~\cite{tran2018closer} developed the R(2+1)D model, dividing a standard convolution into spatial ($M_i$ $1 \times d\times d$) and temporal ($1$ $d\times1\times1$) convolutions, enhancing temporal analysis and network trainability. Xie \textit{et al.} ~\cite{xie2018rethinking} presented the S3D model, which integrates 2D and 3D convolutions using a Top-Heavy structure for improved prediction accuracy by separating 3D convolutions into spatial and temporal parts.

To address the convolution kernel's limited receptive field, Wang \textit{et al.} ~\cite{wang2018non} introduced a Non-local module for global temporal feature extraction. Inspired by biological mechanisms, Feichtenhofer \textit{et al.} ~\cite{feichtenhofer2019slowfast} designed the SlowFast network, leveraging residual connections and Non-Local modules for effective video analysis. Further, Feichtenhofer \textit{et al.} ~\cite{feichtenhofer2020x3d} proposed the X3D model, optimizing computational efficiency and parameter count through machine learning-inspired feature selection.

\subsection{Transformer-based Neural Network}




\subsubsection{Transformer-based Architectures and Spatiotemporal Attention Design}


Transformer-based architectures have been progressively applied to texts, images, and videos, with significant achievements. In the context of videos, designing space-time attention modules is a critical aspect of vision transformer architectures, as illustrated in Fig \ref{network}(c). Recent advancements in transformer-based video action recognition architectures introduce four primary approaches to spatiotemporal attention: split, factorized, joint, and redesigned.

The split approach, as seen in VTN ~\cite{neimark2021video}, utilizes a spatial backbone, a temporal attention encoder, and a classification head, focusing on separating spatial and temporal processing. STAM, proposed by Sharir \textit{et al.} ~\cite{sharir2021image}, follows a similar approach by integrating a temporal aggregating transformer with spatial transformers. In contrast, Arnab \textit{et al.} ~\cite{arnab2021vivit} explored different attention mechanisms and advocated for a \textit{factorized encoder} in ViViT to efficiently combine spatial and temporal information, addressing the computational challenges of \textit{joint space-time attention}.


Joint spatiotemporal attention is implemented in the Video Swin Transformer by Liu \textit{et al.} ~\cite{liu2021video}, which utilizes \textit{shifted windows} ~\cite{liu2021swin} in its self-attention module to achieve a balance between speed and accuracy, demonstrating strong performance across various recognition tasks.

Additional attention mechanisms include \textit{Trajectory Attention} by Patrick \textit{et al.} ~\cite{patrick2021keeping} in the Motionformer, which aggregates information along motion paths, and \textit{Space-time Mixing Attention} by Bulat \textit{et al.} ~\cite{bulat2021space}, enhancing the efficiency of ViT with minimal computational cost.

\subsubsection{Multiscale and Multiview Transformers}

Drawing from the pyramid concept in multiscale processing to simplify their approach, researchers developed similar options for vision transformers. Fig. \ref{network}(d) highlights Fan \textit{et al.}'s~\cite{fan2021multiscale} blend of transformers with CNN's multiscale features, leading to MViT. This model introduces a hierarchical structure with Multi Head Pooling Attention for adaptable resolution handling, expanding channel capacity, and lowering spatial resolution in stages to build a multiscale feature pyramid. Wu \textit{et al.}~\cite{wu2022memvit} further evolved this with MeMViT, enhancing long-term video recognition by utilizing previously stored memory. These advancements are adaptable to various transformer models, according to the researchers. Yan \textit{et al.}~\cite{yan2022multiview} diverged by developing a Multiview Transformer, which processes different video views through distinct encoders, achieving optimal results with cross-view attention. This approach effectively balances accuracy and computational demands for ViT variants. VideoMAE V2~\cite{Wang_2023_CVPR} achieves state-of-the-art performance on various downstream tasks through a designed masking strategy. Hiera~\cite{ryali2023hiera} innovatively achieves a simpler and more efficient hierarchical vision transformer model that excels across multiple visual tasks, by leveraging strong visual pretext task (MAE) pretraining to eliminate complex components traditionally used in hierarchical vision transformers.

\subsubsection{Integration of Transformer and CNN}

Before the emergence of pure-transformer models in computer vision, initial efforts focused on enhancing CNNs with self-attention to improve long-range dependency modeling. 
NLNet, as mentioned by Liu et al., was a forerunner in adding self-attention to CNNs for pixel-level long-range dependencies. Following NLNet, various efforts sought to refine this by simplifying or completely overhauling the non-local block. 
For instance, Cao et al. enhanced NLNet's global context capture with a query-independent approach. With increasing interest in combining Transformers and CNNs, new methods emerged. Xiao et al. showed that adding a convolutional stem to image transformers improves optimization stability and performance without losing computational efficiency. This encourages using early convolution layers in ViT architectures for video analysis. 
Li et al. introduced Uniformer~\cite{li2022uniformer}, a technique for learning spatiotemporal representations that balances global dependency capture and local redundancy reduction in ViT and CNN models. 

\section{Comparison}

\begin{table}[t]
\caption{Comparison of different network architectures.}
\centering
\small
\setlength{\tabcolsep}{1mm}
\resizebox{\columnwidth}{!}{
\begin{tabular}{|l|l|l|l|l|l|}
\hline
\textbf{Two-Stream Networks} & Year & Params (M) & HMDB51 & K400 & SSv2 \\
\hline
TN~\cite{simonyan2014two} & 2014 & 36.6 & 59.4 & - & - \\
TDD~\cite{Wang_2015_CVPR} & 2015 & - & 63.2 & - & - \\
TSN~\cite{wang2016temporal} & 2016 & - & 68.5 & 73.9 & - \\
DOVF~\cite{lan2017deep} & 2017 & - & 71.7 & - & - \\
TLE~\cite{diba2017deep} & 2017 & - & 71.7 & - & - \\
ActionVLED~\cite{girdhar2017actionvlad} & 2017 & - & 66.9 & - & - \\
TSM~\cite{lin2019tsm} & 2019 & 24.3 & 73.5 & 74.1 & 61.7 \\
\hline\hline
\textbf{3D CNNs} & Year & Params (M) & HMDB51 & K400 & SSv2 \\
\hline
C3D~\cite{tran2015learning} & 2015 & 34.8 & 56.8 & 59.5 & - \\
I3D~\cite{carreira2017quo} & 2017 & - & 74.8 & 71.1 & - \\
Two-Stream I3D~\cite{carreira2017quo} & 2017 & 25.0 & 80.9 & 74.2 & - \\
P3D~\cite{qiu2017learning} & 2017 & 98.0 & - & 71.6 & - \\
ResNet3D~\cite{hara2018can} & 2018 & - & 70.2 & 65.1 & - \\
NL I3D~\cite{wang2018non} & 2018 & 61.0 & - & 77.7 & - \\
R(2+1)D~\cite{tran2018closer} & 2018 & 118.2 & 74.5 & 72.0 & - \\
S3D~\cite{xie2018rethinking} & 2018 & 8.8 & 75.9 & 74.7 & - \\
SlowFast~\cite{feichtenhofer2019slowfast} & 2019 & 62.8 & - & 79.8 & - \\
X3D~\cite{feichtenhofer2020x3d} & 2020 & 20.3 & - & 80.4 & - \\
\hline\hline
\textbf{Transformer-based} & Year & Params (M) & HMDB51 & K400 & SSv2 \\
\hline
VTN~\cite{neimark2021video} & 2021 & 114.0 & - & 79.8 & - \\
TimeSFormer~\cite{bertasius2021space} & 2021 & 121.4 & - & 80.7 & 62.4 \\
STAM~\cite{sharir2021image} & 2021 & 96 & - & 80.5 & - \\
ViViT-L~\cite{arnab2021vivit} & 2021 & - & - & 81.7 & 65.9 \\
MViT-B~\cite{fan2021multiscale} & 2021 & 36.6 & - & 81.2 & 67.7 \\
Motionformer~\cite{patrick2021keeping} & 2021 & - & - & 81.1 & 67.1 \\
X-ViT~\cite{bulat2021space} & 2021 & 92.0 & - & 80.2 & 67.2 \\
Swin-L~\cite{liu2021video} & 2021 & 200.0 & - & 84.9 & - \\
UniFormer~\cite{li2022uniformer} & 2022 & 49.8 & - & 83 & 71.2 \\
MTV-B~\cite{yan2022multiview} & 2022 & 310.0 & - & 89.9 & 68.5 \\
MVD~\cite{wang2023masked} & 2022 & 87.8 & - & 87.2 & \textbf{77.3} \\
InternVideo~\cite{wang2022internvideo} & 2022 & 1000.0 & - & \textbf{91.1} & 77.2 \\
Side4Video~\cite{yao2023side4video} & 2023 & 4.0 & - & 88.6 & 75.5 \\
Hiera~\cite{ryali2023hiera} & 2023 & 673.4 & - & 87.8 & 76.5 \\
VideoMAE V2~\cite{Wang_2023_CVPR} & 2023 & 1011.0 & \textbf{88.1} & 90.0 & 77.0 \\
\hline
\end{tabular}}
\label{Result}
\end{table}

We compare various methods and discuss different network backbones and benchmarks. For backbones, we start by comparing models with the same base encoder before collectively evaluating all models. For benchmarks, we select HMDB-51~\cite{kuehne2011hmdb}, Kinetics-400~\cite{kay2017kinetics}, and Something-Something V2~\cite{goyal2017something} due to their widespread use.

The results shown in Tab.~\ref{Result} demonstrate significant progress in the field of video action recognition over the last decade, thanks to both advancements in network architecture design and the introduction of larger datasets. While VideoMAE V2~\cite{Wang_2023_CVPR} achieves outstanding performance, particularly setting new benchmarks on the HMDB-51 datasets with the highest accuracies, it is essential to note that within the Kinetics-400 and Something-Something V2 datasets, other models exhibit leading performance. Specifically, InternVideo~\cite{wang2022internvideo} shows exceptional results on Kinetics-400, and MVD~\cite{wang2023masked} achieves remarkable accuracy on Something-Something V2, highlighting the effectiveness of transformer architectures and their adaptations for video understanding.

Transformer-based methods have generally outperformed two-stream networks and 3D CNNs, showcasing the advantage of leveraging extensive parameters and innovative spatiotemporal attention mechanisms. Among these transformer-based approaches, models like Swin-L~\cite{liu2021video}, UniFormer~\cite{li2022uniformer}, and MVD~\cite{wang2023masked} not only demonstrate the power of joint spatiotemporal processing but also underscore the evolving landscape of video action recognition, where large-scale datasets and complex architectures drive improvements in model accuracy and efficiency. 

Although these methods have reached the pinnacle of accuracy in behavior recognition, considering the requirements for real-time performance in the metaverse, a balance between accuracy and number of parameters needs to be struck in practical applications. For example, the Side4Video~\cite{yao2023side4video} method can ensure accuracy (i.e., 88.6 on K400 and 75.5 on SSv2) while having a smaller number of parameters (4M).

\section{Conclusion}

In reviewing deep neural networks for video action recognition, we explored three primary methods: Two-stream networks, 3D Convolutional Neural Networks (CNNs), and Transformer-based approaches. This survey underscores the crucial role of spatial-temporal information across all methods. 
The two-stream networks combine spatial-temporal information in videos by separately processing temporal and spatial information through designed RGB and optical flow streams. 
The 3D CNN networks consider videos as three-dimensional matrices and then use three-dimensional convolution kernels to fuse the spatial and temporal information.
The transformer-based networks decompose video matrices into different small patches and then fuse the spatial and temporal information within them through attention blocks. 
Although these methods have achieved certain effects in video action recognition in real-world scenarios, when applied to metaverse scenarios, they require new, large datasets for training. At the same time, the computation of the models needs to be further optimized to ensure the real-time requirements of the metaverse. We hope our review will provide a clear and objective reference for future explorations.

\section{Acknowledgement}
This work was supported by National Natural Science Foundation
of China (No. 62203476), Natural Science Foundation
of Shenzhen (No. JCYJ20230807120801002).

\bibliographystyle{IEEEbib}
\bibliography{icme2023template}

\end{document}